\documentclass[letterpaper, 10 pt, conference]{ieeeconf}  

\IEEEoverridecommandlockouts                              

\usepackage{times}
\usepackage{epsfig}
\usepackage{graphicx}
\usepackage{amsmath}
\usepackage{amssymb}
\usepackage{comment}
\usepackage{footnote}
\usepackage{epstopdf}
\usepackage{comment}
\usepackage{multirow}
\usepackage{algorithm}
\usepackage{times}
\usepackage{epsfig}
\usepackage{graphicx}
\usepackage{amsmath}
\usepackage{amssymb}
\usepackage{wasysym}
\usepackage[ruled,vlined,algo2e]{algorithm2e}

\usepackage{color}
\usepackage{dsfont}	

\newcommand{\bfsection}[1]{\vspace*{0.1cm}\noindent\textbf{#1.}}

\title{\LARGE \bf
A Divide-and-Merge Point Cloud Clustering Algorithm for LiDAR Panoptic Segmentation
}

\author{Yiming Zhao, Xiao Zhang, and Xinming Huang
\thanks{Authors are with Department of Electrical and Computer Engineering, Worcester Polytechnic Institute,
        Massachusetts 01609, USA.
        yzhao7@wpi.edu}%
}

\begin{document}

\maketitle
\thispagestyle{empty}
\pagestyle{empty}

\begin{abstract}

Clustering objects from the LiDAR point cloud is an important research problem with many applications such as autonomous driving. To meet the real-time requirement, existing research proposed to apply the connected-component-labeling (CCL) technique on LiDAR spherical range image with a heuristic condition to check if two neighbor points are connected. However, LiDAR range image is different from a binary image which has a deterministic condition to tell if two pixels belong to the same component. The heuristic condition used on the LiDAR range image only works empirically, which suggests the LiDAR clustering algorithm should be robust to potential failures of the empirical heuristic condition. To overcome this challenge, this paper proposes a divide-and-merge LiDAR clustering algorithm. This algorithm firstly conducts clustering in each evenly divided local region, then merges the local clustered small components by voting on edge point pairs. Assuming there are $N$ LiDAR points of objects in total with $m$ divided local regions, the time complexity of the proposed algorithm is $O(N)+O(m^2)$.  A smaller $m$ means the voting will involve more neighbor points, but the time complexity will become larger. So the $m$ controls the trade-off between the time complexity and the clustering accuracy. A proper $m$ helps the proposed algorithm work in real-time as well as maintain good performance. We evaluate the divide-and-merge clustering algorithm on the SemanticKITTI panoptic segmentation benchmark by cascading it with a state-of-the-art semantic segmentation model. The final performance evaluated through the leaderboard achieves the best among all published methods. The proposed algorithm is implemented with C++ and wrapped as a python function. It can be easily used with the modern deep learning framework in python. We will release the code under the following link \footnote{https://github.com/placeforyiming/Divide-and-Merge-LiDAR-Panoptic-Cluster}. 

\end{abstract}

\section{INTRODUCTION}

\begin{figure}[t]
    \includegraphics[width=\linewidth, height=3.5cm]{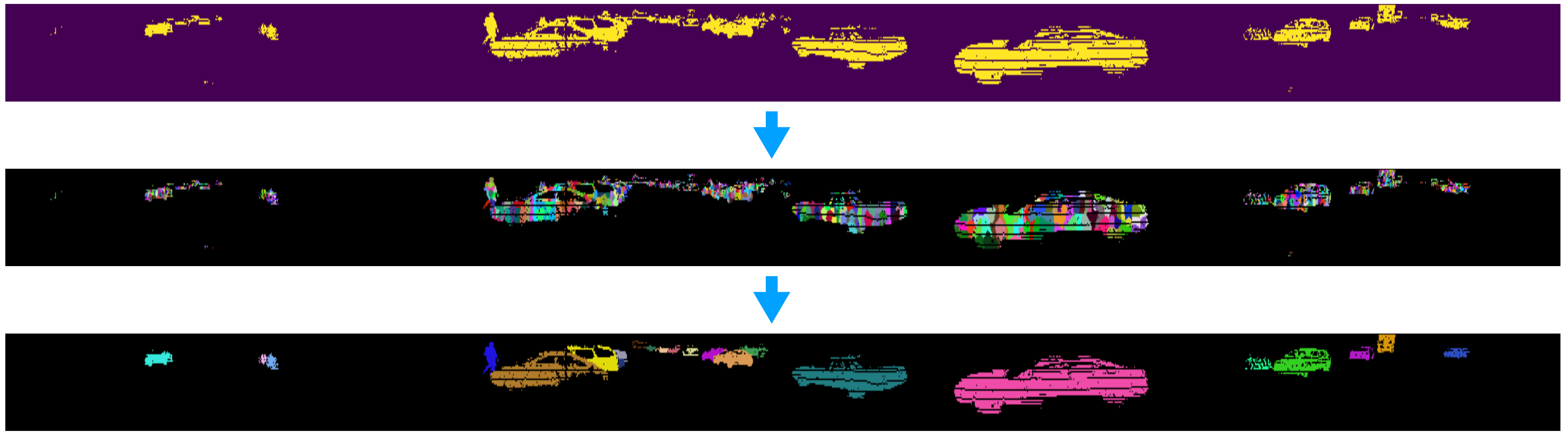}
    \caption{(\textbf{Top}) The object mask generated from a semantic segmentation network. (\textbf{Middle}) The clustering result from each local region. (\textbf{Bottom}) The final instance clustering result after merging local clustered components. This sample is from sequence 19 of the SemanticKITTI test set. The aspect ratio is changed from 64:2048(1:32) to 1:8 for better visualization. }
    \vspace{-0.4cm}
    \label{fig:Cluster_example}
\end{figure}

Point cloud clustering aims to understand the 3D point cloud from a traditional perspective. The research around this area was active before the rise of wide interests in point cloud processing with deep learning methods. \cite{klasing2008clustering, douillard2011segmentation,bogoslavskyi2016fast,douillard2011segmentation}. In our recent paper \cite{zhao2021technical}, we demonstrated combining a  traditional point cloud clustering method \cite{zermas2017fast} with a deep learning semantic segmentation model \cite{zhu2021cylindrical} can achieve the state-of-the-art performance on the SemanticKITTI \cite{behley2019semantickitti} panoptic segmentation benchmark \footnote{https://github.com/placeforyiming/ICCVW21-LiDAR-Panoptic-Segmentation-TradiCV-Survey-of-Point-Cloud-Cluster}. This paper is a follow-up work that we propose a divide-and-merge algorithm for point cloud clustering and evaluate the proposed solution on the panoptic segmentation benchmark by following \cite{zhao2021technical}. Fig. \ref{fig:Cluster_example} illustrates our divide-and-merge pipeline for LiDAR panoptic segmentation. An object mask is generated from a semantic segmentation network first. Then we divide the 3D space into many local regions and perform point cloud clustering in each local region. In the last merging step, the proposed algorithm decides whether to combine neighboring local cluster results by voting on the boundary point pairs.

\bfsection{Motivation} The key of the traditional point cloud clustering method is the condition used to justify if two points belong to the same instance. Euclidean distance with a threshold is a natural choice. But it will group all close objects together as one instance. The angle threshold designed in \cite{bogoslavskyi2016fast} is a heuristic condition that may compensate for the drawback of the naive Euclidean distance. However, this empirical condition can not guarantee it will always work on all point pairs. Considering this, our divide-and-merge algorithm clusters the point cloud into many local components first, then applies this heuristic empirical condition on all edge points and decides whether to merge two components by voting. The original connected-component-labeling (CCL) algorithm used in \cite{bogoslavskyi2016fast} runs fast with linear time complexity $O(N)$, where $N$ is the number of all points. The complexity of the proposed new method has one extra term $O(m^{2})$, where $m$ is the total number of the initial local components. Since $m \ll N$, choosing a proper $m$ will still keep the method working fast and help it maintain good performance as well.

\bfsection{Evaluation for Panoptic Segmentation} Following the same settings of our previous paper \cite{zhao2021surface}, we use the result of a semantic segmentation model Cylnder3D \cite{zhu2021cylindrical} to generate the object mask, then evaluate the proposed cluster algorithm on the point cloud panoptic segmentation benchmark of the SemanticKITTI dataset. We further compare the new method with an existing clustering method called Scan-line Run \cite{zermas2017fast} as well as other panoptic segmentation solutions on the leaderboard. Prior to the era of deep learning, a typical way to understand the point cloud consists of three steps \cite{klasing2009realtime}: ground detection, candidate generating with the clustering algorithm, and handcrafted feature extraction for classification. In the deep learning age, modern point cloud semantic segmentation models work well to predict a semantic label for each point \cite{zhu2021cylindrical}. Thus, the new way to understand the point cloud can be re-organized as two steps: a semantic segmentation network model, and a point cloud cluster as a post-processing step for instances. Therefore, from the view of understanding the point cloud, our panoptic segmentation solution is the counterpart of the traditional way in this deep learning age.

\bfsection{Contributions} This paper has two major contributions: a) We propose a new LiDAR point cloud cluster algorithm, named divide-and-merge cluster. b) We provide a panoptic segmentation solution which outperforms all published methods on the SemanticKITTI panoptic leaderboard. Time complexity of the proposed method is controllable. In our implementation, it can achieve as fast as 20ms for each frame, almost the same as the previous clustering algorithms \cite{bogoslavskyi2016fast,zermas2017fast}, but with a much improved performance.

\section{Related Work}

\bfsection{Clustering with Euclidean Distance} Euclidean cluster firstly constructs kd-tree in the 3D space, then queries neighboring points and groups all points within a threshold as one instance \cite{rusu20113d}. Though sharing the same basic idea, each existing paper has its unique considerations. In \cite{klasing2008clustering}, authors developed a radially bounded nearest neighbor (RBNN) algorithm. Then, they further extended RBNN by involving the normal vector \cite{klasing2009realtime}. This makes the algorithm prone to segment surfaces. In \cite{held2016probabilistic}, a probabilistic framework was proposed to incorporate the temporal information from consecutive frames along with the spatial distance information. For the Euclidean cluster, if the threshold is too large, it would merge close objects together; if the threshold is too small, an object would be split into parts. This under-segmentation and over-segmentation phenomena were discussed in \cite{held2016probabilistic}.

\bfsection{Clustering with Supervoxel or Superpoint} For point cloud processing, supervoxel or superpoint is conceptually similar to superpixel as in image processing. In \cite{papon2013voxel}, authors proposed a voxel cloud connectivity segmentation (VCCS) method which extends the definition of distance used in the iterative image pixel cluster SLIC \cite{achanta2012slic}. From the view of clustering, supervoxel or superpoint usually over segments objects as discussed in \cite{ben2018graph,landrieu2019point}.

\bfsection{Clustering on Range Image} Both the Euclidean cluster and the supervoxel cluster work directly in the 3D world, thus suffer the heavy time cost to query every point. Usually, the point cloud needs to be downsampled first. The other choice to fast cluster point cloud is working on the spherical range image. This gives it a convenience to borrow ideas from image processing techniques. Connected-component labeling (CCL) algorithms\cite{he2017connected,he2008run,wu2009optimizing} are usually used for the LiDAR clustering on range image. One can define a heuristic empirical condition to justify if two neighbor points belong to the same object, such as the distance threshold condition defined in \cite{zermas2017fast} or the angle threshold condition defined in \cite{bogoslavskyi2016fast}. The connected-component labeling (CCL) is a graph algorithm used in computer vision to detect connected regions in binary digital images \cite{he2017connected}. Due to the inherent difference between binary images and LiDAR range images, some modifications need to be done compared with original two-pass CCL \cite{wu2009optimizing} or run-based CCL \cite{he2008run}. The spherical range image clustering algorithms can run in millisecond-level that meet the real-time requirement. This paper's method also works on range image.

\bfsection{LiDAR Panoptic Segmentation} LiDAR panoptic segmentation is a newly defined challenge \cite{behley2020benchmark} inspired by the panoptic segmentation task on the 2D image \cite{kirillov2019panoptic}. The solution of this challenge needs to provide the semantic label for each point and also group points that belong to the same instance together \cite{milioto2020lidar}. The range image representation gives the convenience of directly modifying image-based methods on the point cloud, including both the one-stage DeepLab style method \cite{gasperini2021panoster} and two-stage Mask R-CNN style method \cite{sirohi2021efficientlps}. By considering specific 3D information encoded in the point cloud, it inspired some specially designed network structures, such as a dynamic shifting module developed in \cite{hong2021lidar} or the polar representation \cite{zhou2021panoptic}.

\section{Method}

\begin{figure}[t]
    \includegraphics[width=\linewidth, height=7.cm]{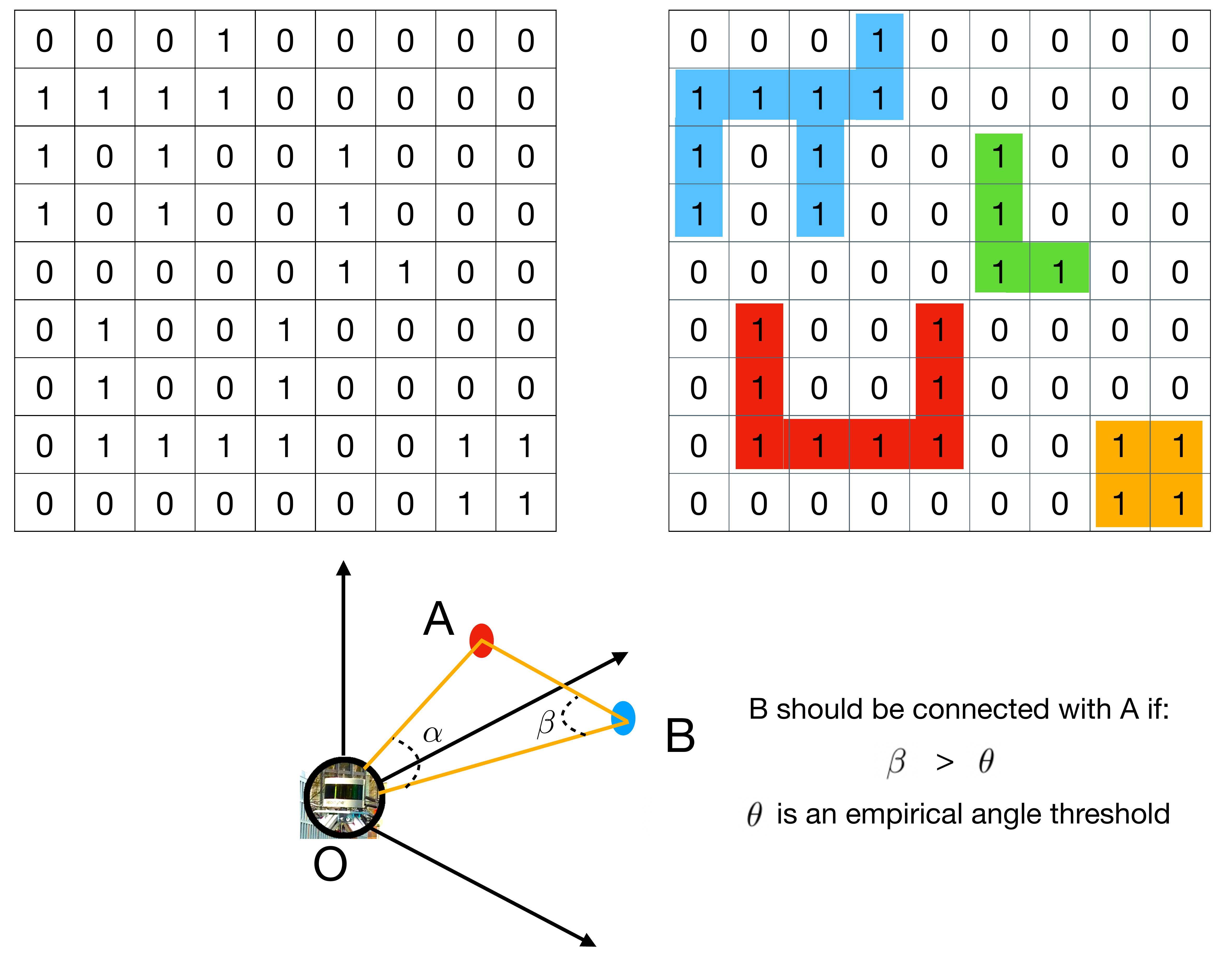}
    \caption{(\textbf{Top})An example about how the connected-component-labeling work on the binary image with a connected condition defined as if two pixels which share the same edge both have value 1. (\textbf{Bottom}) How the depth cluster \cite{bogoslavskyi2016fast} define the heuristic condition to check if two neighbor points come from the same object. }
    \vspace{-0.4cm}
    \label{fig:ccl_example}
\end{figure}

\subsection{Preliminary}

\bfsection{Connected-Component-Labeling on the Binary Image} Connected-component-labeling (CCL) is an algorithmic application of graph theory. The goal of this algorithm is to segment components by checking the pre-defined connected condition. On the top of Fig. \ref{fig:ccl_example}, we show an example of what does the CCL expected to do on a binary image. As a traditional image processing technique, various types of CCL solutions were proposed in the literature, mainly summarized as one-pass algorithms \cite{vincent1991watersheds,abubaker2007one} and two-pass algorithms \cite{he2008run}. Almost all those CCL methods are working in linear complexity with respect to the total number of pixels. There may be a scale factor, but the CLL algorithms run very fast in general. 

\begin{algorithm}[H]\footnotesize
	\caption{Local Clustering}\label{alg:local_cluster}
	\SetAlgoLined
	\SetKwInOut{Input}{Input}\SetKwInOut{Output}{Output}
    \Input{Range image $R$, list of $m$ seeds $L$}
	\Output{Label image $I$, voting matrix $V^{+}$ and $V^{-}$}
	\BlankLine
	
	\hspace{0.2cm}$I \leftarrow$ $zeros(R_{rows}\times R_{cols})$;\\
	$V^{+} \leftarrow$ $zeros(m \times m)$;\\
	$V^{-} \leftarrow$ $zeros(m \times m)$;\\
	$Q \leftarrow$ an empty list of queue;\\
	$undecided\_list \leftarrow$ [];\\
    \For{$i$ \textbf{in} $[0,m)$ }{
    \hspace{0.2cm}$each\_queue \leftarrow$ $queue()$;\\
    $\{r,c\}=L[i]$;\\
    $each\_queue.push(\{r,c\})$;\\
    $Q.push\_back(each\_queue)$;\\
    $I_{\{r,c\}}=i+1$;
       }
      \While{$Q.size()<m$}{
       \For{$i$ \textbf{in} $[0, Q.size())$ }{
       $\textbf{ExpandCluster}(Q[i],R,I,V^{+},V^{-})$;\\
       \If{$Q[i].is\_empty()$}{$Q.delete(Q[i])$}
       }
      }
      \For{$\{r,c\},\{r_n,c_n\}$ \textbf{in} $undecided\_list$ }{
      $V^{-}[I_{\{r_n,c_n\}},I_{\{r,c\}}]+=1$;}
      
     \Return{$I,V^{+},V^{-}$}
     \\
    \setcounter{AlgoLine}{0}
    \SetKwProg{myproc}{Procedure}{\hspace{0.1cm}$\textbf{ExpandCluster}(Q[i],R,I,V^{+},V^{-})$}{end}
    \myproc{}{
        $current\_point=Q[i].front()$;\\
        $Q[i].pop()$;\\
        \For {$\{r_n,c_n\} \in \textbf{Neighborhood}\{r,c\}$}{
        \If{$\textbf{Condition}(\{r,c\},\{r_n,c_n\})$ and $I_{\{r_n,c_n\}}==0$ }{
        \hspace{0.2cm}$Q[i].push(\{r,c\})$;\\
        $I_{\{r_n,c_n\}}=I_{\{r,c\}}$;\\
        }
        \If{$\textbf{Condition}(\{r,c\},\{r_n,c_n\})$ and $I_{\{r_n,c_n\}}>0$ }{
        \hspace{0.2cm}$V^{+}[I_{\{r,c\}},I_{\{r_n,c_n\}}]+=1$;\\
        $V^{+}[I_{\{r_n,c_n\}},I_{\{r,c\}}]+=1$;\\
        }
        \If{(not $\textbf{Condition}(\{r,c\},\{r_n,c_n\})$) and $I_{\{r_n,c_n\}}>0$ }{
        \hspace{0.2cm}$V^{-}[I_{\{r,c\}},I_{\{r_n,c_n\}}]+=1$;\\
        $V^{-}[I_{\{r_n,c_n\}},I_{\{r,c\}}]+=1$;\\
        }
        \If{(not $\textbf{Condition}(\{r,c\},\{r_n,c_n\})$) and $I_{\{r_n,c_n\}}==0$ }{
        $undecided\_list.push\_back([\{r,c\},\{r_n,c_n\}])$;
        }
        }
    }
\end{algorithm} 

\bfsection{Depth Cluster \cite{bogoslavskyi2016fast}: an Example of How CCL Works on LiDAR Range Image} Projecting the point cloud on range image is a commonly used technique to solve LiDAR related challenges \cite{milioto2019rangenet++,zhao2021surface}. However, unlike a binary image with a clear condition to determine if two pixels belong to the same component, it is challenging to define the similar condition on the range image where two neighbor pixels are representing two LiDAR points instead of a binary value. An Euclidean distance threshold can be used as the condition, but it apparently can not distinguish two close objects. In \cite{bogoslavskyi2016fast}, an angle threshold is designed as the condition. We demonstrate this threshold at the bottom of the Fig. \ref{fig:ccl_example}. Compared with Euclidean distance, this heuristic angle condition has no limitation for close objects, so we adopt it in our model. We choose the threshold $\theta = 10 ^{o}$ as suggested in the original paper \cite{bogoslavskyi2016fast}.

\subsection{The Divide Step: Clustering Locally}

On a binary image, if CCL connects two pixels, it is deterministic those two pixels belong to the same component. However, the heuristic condition on the LiDAR range image cannot guarantee the same case. To increase the reliability of the CCL on range image, we propose to firstly cluster objects locally, then merge them by considering all the neighboring points on the boundary instead of a single point pair.

\begin{algorithm}[H]\footnotesize
	\caption{Voting\&Merging}\label{alg:vote_merge}
	\SetAlgoLined
	\SetKwInOut{Input}{Input}\SetKwInOut{Output}{Output}
    \Input{ $V^{+}$, $V^{-}$, label image $I$}
	\Output{Label image $I$}
	\BlankLine
	
	\hspace{0.2cm}$label\_list \leftarrow$ [0, 1, 2, 3, ..., m-1];\\
	 $merged\_label\_list\leftarrow $ [0, 1, 2, 3, ..., m-1];\\
	 $current\_label \leftarrow$ 0;\\
	 \While{$label\_list.size()>0$}{
	  $current\_label$+=1;\\
	  $label\_queue \leftarrow queue()$;\\
	  $label\_queue.push(label\_list[0])$;\\ 
	  $merged\_label\_list[label\_list[0]]=current\_label$;\\
	  
	  \While{$label\_queue.size()>0$}{
	  $target\_label \leftarrow label\_queue.front()$;\\
	  $label\_queue.pop()$;\\
	  $label\_list.delete(target\_label)$;\\
	  \For{$query\_label$ \textbf{in} $label\_list$}{
	  \hspace{0.2cm}$pos \leftarrow V^{+}[target\_label,query\_label]$;\\
	  $neg \leftarrow V^{-}[target\_label,query\_label]$;\\
	  \If{$pos>neg$}{
	  \hspace{0.2cm}$label\_queue.push(query\_label)$;\\
	  $merged\_label\_list[query\_label]=current\_label$;\\
	  \For{$i$ \textbf{in} $[0,m)$}{
    \hspace{0.2cm}$V^{+}[query\_label,i]+=V^{+}[target\_label,i]$;\\
	  $V^{-}[query\_label,i]+=V^{-}[target\_label,i]$;\\
	  }
	  }
	  }
	  
	  }
	 }
	 \For{$r=1,...,I_{rows}$}{
     \For{$c=1,...,I_{cols}$}{
     \If{$I_{\{r,c\}}>0$}{$I_{\{r,c\}}=merged\_label\_list[I_{\{r,c\}}-1]$;\\}
     }
     }
     \Return{$I$}
\end{algorithm}

\begin{figure*}[t]
    \includegraphics[width=\linewidth, height=4.5cm]{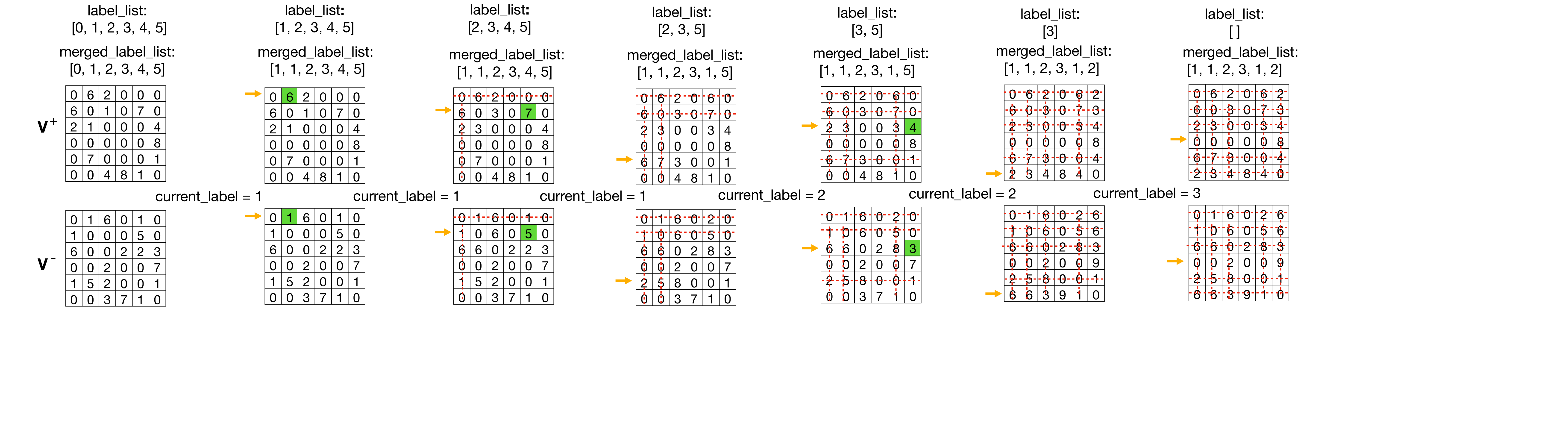}
    \caption{Visualization of an example about how the merge step work with $m=6$. The yellow arrow always points on the row of target label, and the column of the green marker indicates which query label should be merged.} 
    \vspace{-0.cm}
    \label{fig:merge_example}
\end{figure*}

We divide the 3D world as $l \times l \times l$ voxels and pick one LiDAR point in each voxel as the initial seed. Those voxels that do not contain any points will be ignored. Assuming there are $m$ voxels with inside points, the initial number of components is $m$. During the local clustering, no new label will be created. To record the boundary information while clustering, we define two $m \times m$ voting matrix $V^{+}$ and $V^{-}$ , which record the number of point pairs that satisfy or do not satisfy the heuristic condition respectively. The whole algorithm of the local clustering is presented in Alg. \ref{alg:local_cluster}. The \textbf{Neighborhood$\{r,c\}$} is a function to find the 4-connected neighborhood of the pixel location $\{r,c\}$ on the range image. The \textbf{Condition($\cdot,\cdot$)} can be any heuristic condition to determine if two points should be counted as one component. Here we use the same angle threshold condition proposed in \cite{bogoslavskyi2016fast}. This condition has been explained in Fig. \ref{fig:ccl_example}.

\subsection{The Merge Step: Voting on the Edge}

After the local clustering, we will have a labeled image with many small components with connected points, as shown in the middle of Fig. \ref{fig:Cluster_example}. The local cluster algorithm will not create new labels, so the total number of different labels is $m$. A point will be repeatedly considered by all components that can reach it instead of directly creating a new label as \cite{bogoslavskyi2016fast}. This is the key idea of our method. The second voting and merging step will only work on $V^{+}$, $V^{-}$, and the label image $I$. We present the detailed process about how to vote and merge all components in Alg. \ref{alg:vote_merge}. To better explain the merge step, we exhibit a simple example in Fig. \ref{fig:merge_example} with $m=6$. 

\subsection{Time Complexity} 

In the divide step, those points that are not on the edge of two local components will only be visited once. Those points on the edge of local components may be visited multiple times, but only limited number of voxels have chance to reach them. So the time complexity of the divide step is linearly proportional to the total number of points $N$. In the merge step, each label will be visited once, but the edge information needs to be updated for each merging. Thus the time complexity of the merge step is proportional to the square of the total number of labels $m$. In summary, the time complexity of our proposed divide-and-merge pipeline is linearly proportional to $O(N)+O(m^2)$. For the KITTI dataset, each frame contains around 100k points. The $N$ is depending on how many points of objects are in the frame, which is around 10k points. If we set the size of each local voxel as $l=1m$, the $m$ is around 200, then the merge step has a similar time complexity as the divide step. However, if the local voxel is small, like $l=0.2m$, the $m$ will be around $1k$, then the merge step will dominant the time cost. We give a numerical comparison with our implementation in Section IV.

\subsection{Post-process on KITTI} 

\begin{figure}[t]
    \includegraphics[width=\linewidth, height=2.5cm]{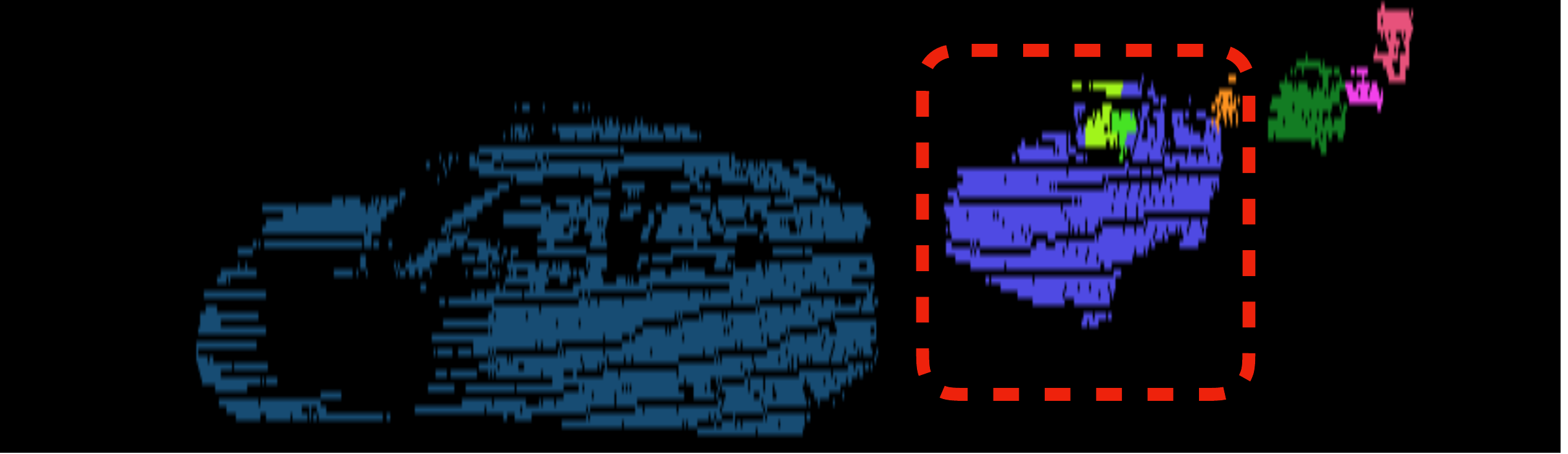}
    \caption{Illustrate the motivation of the post-process step on KITTI. Drivers inside the car are clustered as separate objects, whereas the panoptic benchmark on SemanticKITTI recognizes both the driver and the car as one single object.} 
    \vspace{-0.4cm}
    \label{fig:post_example}
\end{figure}

What is the definition of an object? This question sounds trivial but needs to be considered for better performance on the panoptic segmentation benchmark of the SemanticKITTI dataset. In this deep learning age, models are usually trained with labels, the label implies the definition of objects. However, as the traditional algorithms, point cloud cluster methods do not have an explicit definition of the object. We show an clustered example in Fig. \ref{fig:post_example}. LiDAR points flight through the windshield, and are reflected back from the driver sitting inside a car. Our cluster method separates driver from the car, but the panoptic benchmark of the SemanticKITTI dataset recognizes them as one single object. Thus we define a post-process step that two components need to be merged if they share the same semantic label and have an overlapping region on the 2D birds-eye-view plane. Note, this post-process step is purely for coping with the object definition of the SemanticKITTI dataset. One can choose to keep or discard it while using the algorithm on custom dataset or applications.

\section{Experiment}

\begin{table*}
\begin{center}
    \caption{Performance comparison on SemanticKITTI test set (sequence 11 to 21). We use blue to highlight our runner-up indicators. }
        \label{tab:table1}
    \setlength{\tabcolsep}{2.pt}
      \renewcommand{\arraystretch}{1.5} 
 \begin{tabular}{c|c|c|c c c c| c c c| c c c |c } 
  \hline
 Methods & Type & Venue& $\mathbf{PQ}$\hspace{0.1cm} & $\mathbf{PQ^{\dagger}}$ \hspace{0.1cm}& $RQ$\hspace{0.1cm} & $SQ$\hspace{0.1cm} &$PQ^{Th}$ & $RQ^{Th}$ & $SQ^{Th}$ & $PQ^{St}$ & $RQ^{St}$ &  $SQ^{St}$& $mIoU$\\ 
 \hline
 \hline
KPConv \cite{thomas2019kpconv} + PointPillars \cite{lang2019pointpillars}&semantic+3D bbox &-& 44.5& 52.5& 54.4& 80.0 & 32.7& 38.7 &81.5 & 53.1 &65.9 &79.0&58.8\\
 
RangeNet++ \cite{milioto2019rangenet++} + PointPillars \cite{lang2019pointpillars}&semantic+3D bbox&-& 37.1& 45.9& 47.0& 75.9 & 20.2& 25.2 &75.2 & 49.3 &62.8 &76.5&52.4\\

KPConv \cite{thomas2019kpconv} + PV-RCNN \cite{shi2020pv}&semantic+3D bbox &-& 50.2& 57.5& 61.4& 80.0 & 43.2& 51.4 &80.2 & 55.9 &68.7 &79.9&62.8\\

Panoptic RangeNet \cite{milioto2020lidar}&panoptic &IROS'20& 38.0& 47.0& 48.2& 76.5 & 25.6& 31.8 &76.8 & 47.1 &60.1 &76.2&50.9\\

Panoster \cite{gasperini2021panoster}&panoptic &RA-L'21& 52.7& 59.9& 64.1& 80.7 & 49.4& 58.5 &83.3 & 55.1 &68.2 &78.8&59.9\\

PolarNet\_seg \cite{zhou2021panoptic}&panoptic &CVPR'21& 54.1& 60.7& 66.0& 81.4 & 53.3& 60.6 &87.2& 54.8 &68.1 &77.2&59.5\\

DS-Net \cite{hong2021lidar}&panoptic &CVPR'21& 55.9& 62.5& 66.7& 82.3 & \textbf{55.1}& \textbf{62.8} &\textbf{87.2} & 56.5 &69.5 &78.7&61.6\\

Cylinder3D \cite{zhu2021cylindrical} + SLR \cite{zermas2017fast} \cite{zhao2021technical}&semantic+cluster &ICCVW'21& 56.0& 62.6& 67.4& 82.1 & 51.8& 61.0 &84.2 & 59.1 &72.1 &80.6&67.9\\

 \hline
Cylinder3D\cite{zhu2021cylindrical} + Ours&semantic+cluster & - & \textbf{56.5}& \textbf{63.1}& \textbf{67.9}& \textbf{82.3} & 52.9& \textcolor{blue}{62.1} &\textcolor{blue}{84.8} & \textbf{59.1} &\textbf{72.1} &\textbf{80.6}&\textbf{68.2}\\

 \hline
\end{tabular}
\end{center}
    \vspace{-0mm}
\end{table*}

\subsection{Dataset}

SemanticKITTI \cite{behley2019semantickitti} is a point cloud dataset about outdoor autonomous driving. It provides the benchmark of several tasks, including point cloud semantic segmentation and point cloud panoptic segmentation \cite{behley2020benchmark}. The dataset contains a training split with ten LiDAR sequences, a validation split with one sequence, and a test split with eleven sequences. All labels on the test split are unavailable. Users need to submit the prediction results to the leaderboard for final evaluation scores. To better present our clustering method, we choose the best semantic segmentation model with open-sourced code\footnote{https://github.com/xinge008/Cylinder3D} to provide the semantic prior \cite{zhu2021cylindrical}. 

\subsection{Metrics}
The major indicator to measure the panoptic segmentation is $PQ$ defined in \cite{behley2020benchmark}. Those non-object semantic classes, like road and building, can be simply assigned one instance id. Then, $S$ and $\hat{S}$ represent a set of points of one predicted and ground truth segment id respectively, which are unique by considering both the semantic and instance. The $PQ$ of each id is defined as:
\begin{align}\label{hh}
PQ_{c}=\frac{\sum_{(S,\hat{S}) \in TP_{c}} IoU (S, \hat{S})}{|TP_{c}|+\frac{1}{2}|FN_{c}|+\frac{1}{2}|FP_{c}|}
\end{align}
The final PQ is defined as:
\begin{align}
PQ=Average(PQ_{c})
\end{align}
In Eq. \ref{hh}, $IoU(S,\hat{S})=(S\cap \hat{S})\cdot(S \cup \hat{S})^{-1}$, and others are calculated following the same way, such as $TP_{c}=\{ (S,\hat{S})| IoU(S,\hat{S})>0.5\}$. This $PQ$ is argued in favor of instance, so an adjusted $PQ^{\dagger}$ is proposed in \cite{porzi2019seamless}. One can check more details about the $PQ^{\dagger}$ on point cloud with other indicators by referring to \cite{behley2020benchmark}.

Although the $PQ$ considers more about the instance, it is still not solely evaluating the instance clustering. When we compare with other cluster algorithms, we use the same semantic segmentation result to make the comparison fair. When we compare with other panoptic segmentation solutions, this indicator is used to illustrate our clustering pipeline can form a strong panoptic solution by directly applying it to an existing semantic model. 

\subsection{Performance Comparison}
\bfsection{Compare with Cluster Methods} In \cite{zhao2021technical}, we show the Scan-line Run (SLR) \cite{zermas2017fast} is the best one among four clustering methods evaluated on the SemanticKITTI validation set. We compare our divide-and-merge method with the SLR on the test leaderboard with exactly the same semantic predictions from Cylinder3D \cite{zhu2021cylindrical}. In Table \ref{tab:table1}, we can see those indicators designed for the semantic part are the same, but our method outperforms SLR on all indicators about the instance clustering performance.

\begin{figure}[t]
\centering
    \includegraphics[width=0.8\linewidth, height=5.cm]{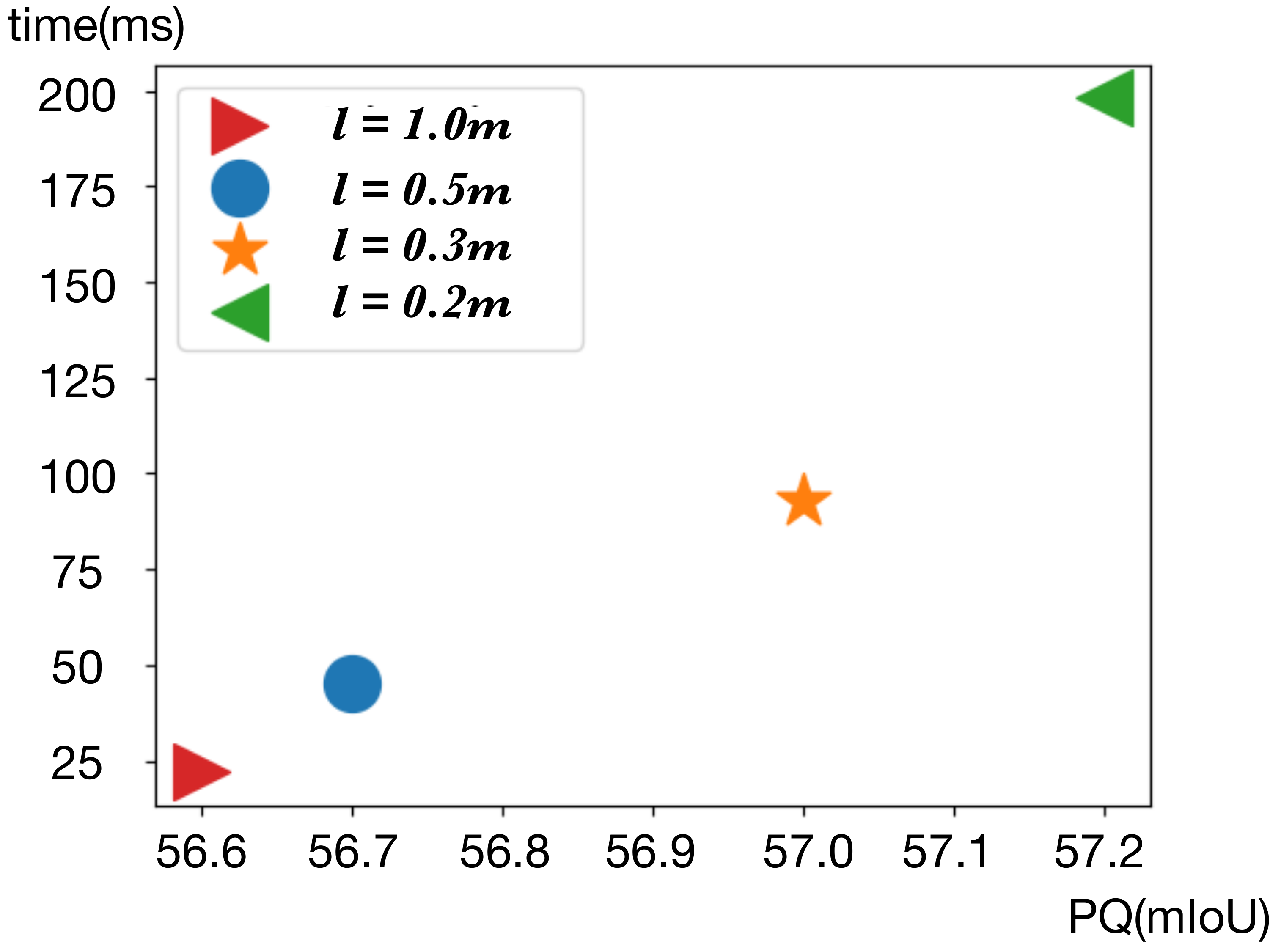}
    \caption{Method performance on the validation sequence 08 with various voxel size $l$. } 
    \vspace{-0.cm}
    \label{fig:tradeoff_example}
\end{figure}

\begin{figure*}[t]
\centering
    \includegraphics[width=1.\linewidth, height=8cm]{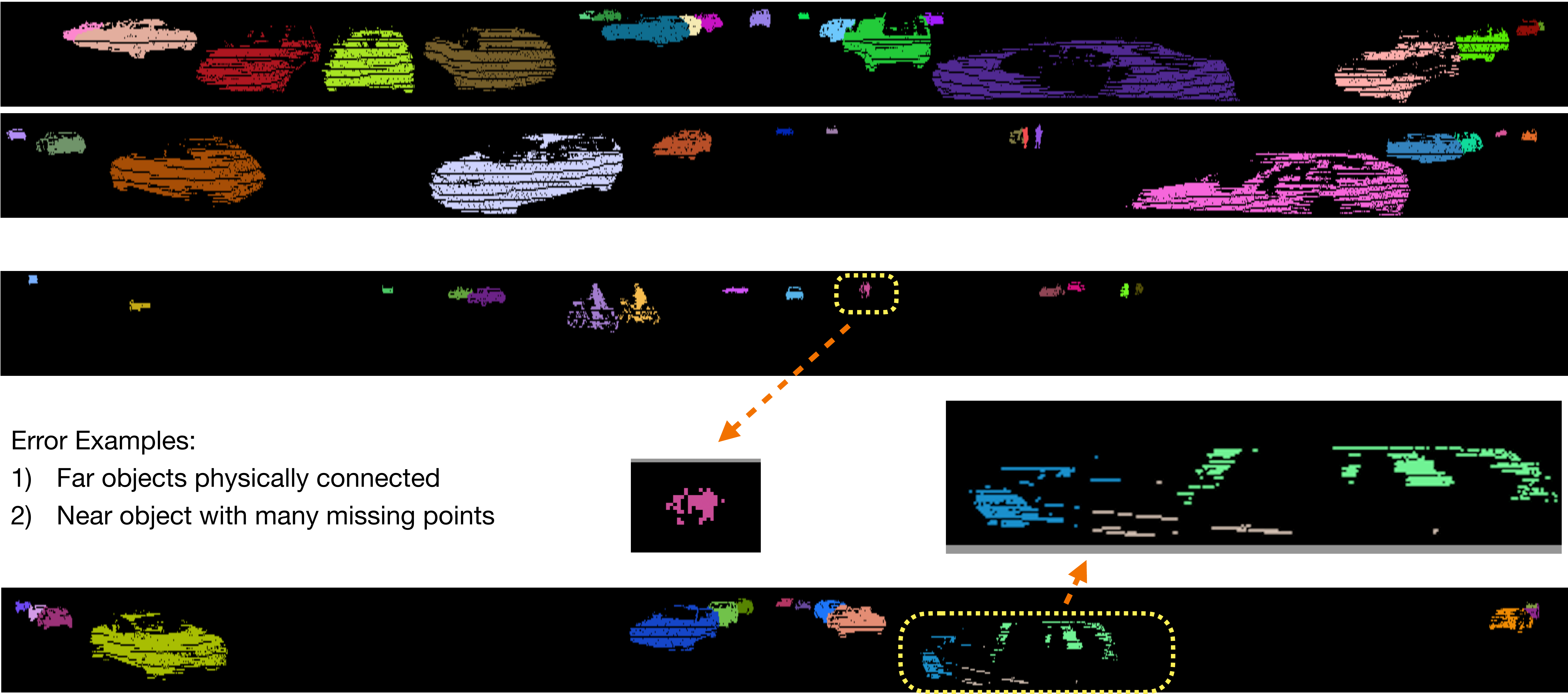}
    \caption{The top two examples show our method works well for most cases, even cars are parking very close. In the bottom two examples, we present two typical error situations: almost physically connected far objects and close object with many missing points. All four LiDAR frames are picked from the sequence 19 in the test set. } 
    \vspace{-0.cm}
    \label{fig:good_bad_example}
\end{figure*}

\bfsection{Compare with Other Panoptic Solutions}
In Table \ref{tab:table1}, we also compare the hybrid panoptic segmentation solution of this paper with other panoptic solutions. Combining semantic segmentation with 3D object detection is also a branch of hybrid solutions. But the 3D detection needs an extra large network with redundant computational cost and labels. Our hybrid solution with the traditional clustering method shows better performance than those hybrid solutions with 3D detection reported in \cite{hong2021lidar}. In terms of those end-to-end panoptic segmentation network models, our pipeline achieves comparable performance with the current state-of-the-art model DS-Net \cite{hong2021lidar}.

\bfsection{Relationship with Semantic Segmentation} By comparing with the DS-Net \cite{hong2021lidar} as in Table \ref{tab:table1}, we show our hybrid solution performs better for $PQ^{St}$, $RQ^{St}$ and $SQ^{St}$, but performs worse for $PQ^{Th}$, $RQ^{Th}$ and $SQ^{Th}$. Those indicators $PQ^{St}$, $RQ^{St}$, and $SQ^{St}$ measure the model performance on non-object points. This phenomenon has been discussed in \cite{zhao2021technical}. It implies the traditional cluster algorithm  keeps the same high performance of the semantic segmentation model while provides very good instance clustering results. This is a benefit of using traditional methods for panoptic segmentation, and explains why this hybrid solution perform well.

\subsection{Time Complexity}

From previous discussions in Section III, we know a smaller local voxel size $l$ means a larger component number $m$, which should give better performance but leads to a larger time cost. So we report the trade-off between the running time and the performance in Fig. \ref{fig:tradeoff_example}. Note, the inference time is the entire time cost of our C++ implementation on the SemanticKITTI dataset, including the post-processing step. We tested inference speed on i7-6700K CPU @ 4.00GHz. The actual speed can be improved further with advanced coding features like semantic move or with multi-threading programming on multiple CPU cores. The divide-and-merge pipeline is inherently suitable for multi-threading programming, but the engineering optimization is out of the scope of this paper. The test result in Table \ref{tab:table1} is achieved by choosing $l=0.5m$.

\section{Discussion}

Here we discuss two typical observed failure situations of our proposed method while we running it on the SemanticKITTI dataset. After visually checking many frames, we realize there are some situations that our clustering method is prone to fail. The first situation is when two objects with large distance to the LiDAR sensor are likely connected with each other in the physical world. The third LiDAR frame in Fig. \ref{fig:good_bad_example} shows a case that two far away pedestrians stay very close and are clustered as one object. The second situation is when cars with reflective surfaces are staying close to the LiDAR sensor. As shown at the bottom of Fig. \ref{fig:good_bad_example}, the LiDAR sensor will miss many laser beams and creates a largely empty patch on the range image. Correctly searching neighbors on this large empty patch is almost impossible. Even our divide-and-merge pipeline cannot overcome this sensor limitation. 

\section{Conclusion}

This paper provides a new point cloud clustering algorithm, named divide-and-merge clustering. It focuses on solving the potential problem of directly using the connected-component-labeling method designed for the binary image on the LiDAR spherical range image. By combining with a well-performing semantic segmentation model, the proposed clustering algorithm also delivers a strong solution for LiDAR panoptic segmentation task. We give an analysis of the time complexity, and also report the real inference time with our implementation. By sharing the code of our implementation, we believe the contribution of this paper is useful for both academic research and many industrial robot applications including perception, localization, and map building.




\bibliographystyle{IEEEtran}
\bibliography{IEEEabrv,IEEEexample}

\end{document}